\documentclass{article}

  \usepackage[dblblindworkshop, final]{neurips_2026}

\workshoptitle{The 6th Workshop on Mathematical Reasoning and AI}

\usepackage[utf8]{inputenc} 
\usepackage[T1]{fontenc}    
\usepackage{hyperref}       
\usepackage{url}            
\usepackage{booktabs}       
\usepackage{amsfonts}       
\usepackage{nicefrac}       
\usepackage{microtype}      
\usepackage{xcolor}         

\usepackage{amsmath}
\usepackage{amssymb}
\usepackage{graphicx}

\newcommand{\vlv}{\textsc{Viva La Vida}}
\newcommand{\gateone}{\textsc{Triage}}
\newcommand{\gatetwo}{\textsc{Verify}}
\newcommand{\gatethree}{\textsc{Extract}}

\title{\vlv{}: Verification and Accumulation Failures in Multi-Agent Proof Search}

\author{%
  Benji Xu \qquad Ken Zheng \qquad Noah Han \\
  University of California, Berkeley \\
  \texttt{\{benji.xu, kenzhengjk, noah0427\}@berkeley.edu}
}

\begin{document}

\maketitle

\begin{abstract}
When an agentic prover works on an open problem, there is no proof assistant to fall back on: its verifier and lemma library are ultimately language models judging model outputs. We instrumented such a system end to end and analyzed $51{,}754$ traced observations across three full runs ($186$ hours, \$$5{,}694$). We find three connected failure modes. First, the three-model verifier requires unanimity and treats parse or API failure as non-approval; in $10$ of $12$ verification events, one member returned no parseable output or an API error, making acceptance arithmetically impossible without surfacing an error. Second, when the ensemble did function, one verifier approved $3$ attempts that GPT rejected, each claiming to resolve the open problem; a single-verifier design would therefore have announced a solution three times. Third, because nothing could be approved, every review was a refutation, yet the lemma extractor mines reviews as well as proofs: $24$ of $93$ lemmas ($26\%$) were extracted from rejected arguments with their refutational context removed. Taken together, these findings show that without external verification, supervision is itself a critical trust boundary: systems must distinguish abstention from rejection, preserve useful disagreement, and preserve the provenance and polarity of information before it becomes future context. 
We release traces and analysis scripts \href{https://github.com/kenzhengjk/viva-la-vida}{here}.
\end{abstract}

\section{Introduction}

Memory-augmented provers are attractive in part because a failed run can still leave behind useful artifacts: a lemma, a reduction, or a ruled-out direction. On benchmark problems, the question of what to keep has a clean answer, since a proof assistant can check each candidate \citep{yang2023leandojo, wang2025kimina, xin2025bfsprover}. On an open problem, this is no longer the case. The relevant mathematics is largely unformalized \citep{leandojo2026millennium}, informal proof is where current models perform best \citep{varambally2025hilbert}, and no oracle can say whether a proposed lemma is actually true. Both verification and retention therefore become judgment calls made by other models.

We built \vlv{}, an agentic prover targeting the exact optimal erasure resilience ratio $\alpha^\star$ for binary interactive error-correcting codes (iECCs), where the known bounds leave a gap \citep{gupta2022interactive}. Recursive prover agents write to a shared library behind three gates (Section~\ref{sec:system}). We ran the system three times, once per prover backbone, and traced every model call. This paper is not about $\alpha^\star$ itself; rather, it is about what the gates did in practice and what ended up in the library as a result.

\textbf{The target problem.} In an interactive error-correcting code, Alice holds a message $x$, and Bob must learn it by exchanging bits with her over a binary channel while an adversary erases part of the communication. The adversary's budget is a fraction of the \text{total} communication, so bits sent in both directions count against it. The erasure resilience ratio $\alpha^\star$ is the largest such fraction that some binary iECC can tolerate while Bob still recovers $x$. Interaction is what makes the question interesting: \citet{gupta2022interactive} show that it lets binary codes survive the erasure of more than half of all communication, constructing an iECC resilient to $3/5$ and proving that no binary iECC is resilient to $2/3$. The exact value in between is open. \vlv{} is built to search for informal proofs of open statements like this one, which is also why no proof assistant can check its output.

\section{\vlv's Tri-Gate Architecture}
\label{sec:system}

\begin{figure}[t]
  \centering
  \includegraphics[width=\linewidth]{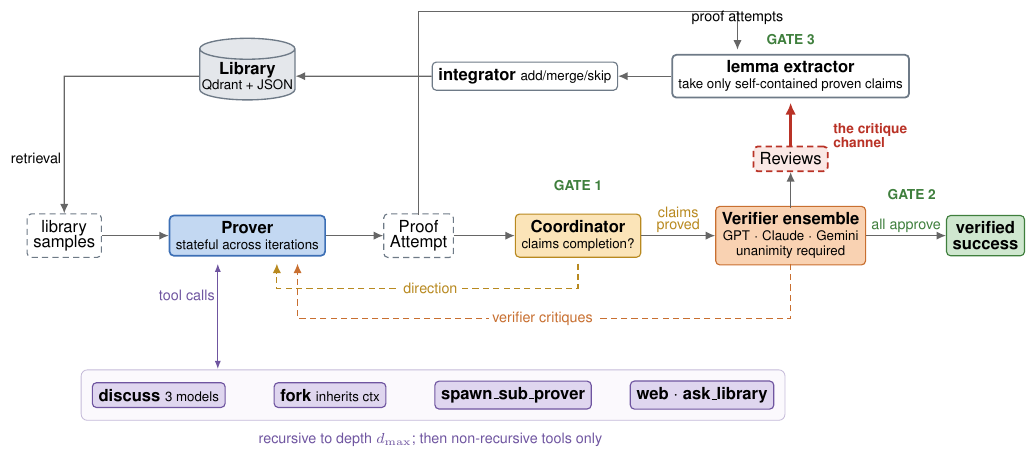}
  \caption{\vlv{}. The spine is one iteration: a stateful prover with recursive
  (\texttt{fork}, \texttt{spawn\_sub\_prover}) and non-recursive tools emits a
  proof attempt; a coordinator triages whether it \text{claims} completion; if so, a
  three-model ensemble must approve unanimously. Above the spine is the intake
  path: proof attempts \text{and reviews} are both mined for lemmas by a
  conservative extractor, curated, and embedded for retrieval by later provers.
  All three gates scrutinize the prover. The red edge is the channel none of them
  was designed for.}
  \label{fig:arch}
\end{figure}

The \gateone{} gate (coordinator) decides only whether an attempt \text{claims} completion, not whether it is correct. The \gatetwo{} gate (verifier ensemble) runs GPT-5.4, Claude Opus 4.7, and Gemini 3.1 Pro in parallel, and all three must approve. The gate is fail-closed: an API error, turn-limit exhaustion, or unparseable response is recorded as a review with \texttt{approved=false}. The \gatethree{} gate (lemma extractor) is deliberately restrictive. It emits only (statement, proof) pairs that the source both states and proves, never repairs, extends, or supplies a missing proof, and returning zero pairs is acceptable. This extractor is applied to every proof attempt \text{and every review}. Survivors are passed to an integrator that may skip, merge, or add them, and are then embedded for retrieval.
See Figure~\ref{alg:pipeline} for a more structured presentation of the \vlv{} algorithm.


\begin{table}[t]
  \centering
  \caption{The three full runs. ``Iters'' counts
  coordinator judgements; ``Verified'' counts attempts reaching the ensemble.}
  \label{tab:runs}
  \small
  \begin{tabular}{lrrrrrrr}
    \toprule
    Prover backbone & Hours & Cost & Generations & Iters & Verified & \texttt{fork} & \texttt{spawn} \\
    \midrule
    GPT-5.4         &  91.3 & \$3{,}517 & 11{,}743 & 10 &  0 & 404 & 689 \\
    Gemini 3.1 Pro  &  81.4 & \$1{,}947 &  4{,}929 & 12 & 10 & 366 &  30 \\
    Claude Opus 4.7 &  13.9 & \$229     &     881  & 15 &  2 &  17 &  40 \\
    \midrule
    Total           & 186.6 & \$5{,}694 & 17{,}553 & 37 & 12 & 787 & 759 \\
    \bottomrule
  \end{tabular}
\end{table}

\section{Result 1: The Ensemble Could Not Approve}

Twelve attempts reached \gatetwo{}, producing $36$ reviews in total, with zero unanimous approvals. That number is misleading on its own, however, because in most cases approval was not actually possible.
In fact, ten of the $36$ reviews were not mathematical judgments at all; all ten came from the Claude verifier. Eight carry the reason string \texttt{"Claude returned no parsed output."}, meaning the model produced no parseable structured response, and two record an API error. The fail-closed path stored all ten as non-approvals, identical to a rejection on the merits. Under the unanimity rule, a single guaranteed non-approval vetoes the entire ensemble. So in $10$ of $12$ verification events ($83\%$), acceptance was arithmetically impossible before any mathematics was read. The pipeline's only success state was unreachable, and nothing surfaced this to the operator: runs completed normally, and the outcome was indistinguishable from a correct unanimous rejection.
An infrastructure failure should not admit an unchecked proof, and unanimity is the conservative aggregation rule. However, conservative-on-error composed with conservative-on-aggregation does not yield a safe verifier but an inert one that always rejects. Separating abstention and rejection as distinct booleans and treating an aggregate containing an abstention as an error that is surfaced in real-time could be an interesting follow-up work here.

\textbf{The same pattern beyond verification.} \vlv{} caps prover recursion at depth $5$, beyond which sub-provers lose \texttt{fork} and \texttt{spawn\_sub\_prover}, and the traces record the depth at every level. This bound held for the GPT backbone, but the Claude backbone reached depth $8$ and the Gemini backbone depth $108$. Like the verifier, the guard was correct on the path we developed against and silently absent on another, and nothing in the run output revealed it.

\begin{figure}[t]
\begin{center}
\fbox{
\begin{minipage}{0.93\linewidth}
\textbf{\vlv{} Algorithm.}

\textbf{Input:} Problem $x$, library $L$, max iterations $T$, max recursion depth $d_{\max}$.
\textbf{Init:} System prompt and \emph{uniform} sample of $k$ artifacts (proofs, reviews, lemmas) from $L$.

\textbf{For} $i=1,\ldots,T$:
\begin{enumerate}
  \item Prover generates attempt $P_i$ via conversation and tools (\texttt{discuss}, \texttt{web}, \texttt{ask\_library}, \texttt{fork}, \texttt{spawn\_sub\_prover} if depth $< d_{\max}$). If empty, save, re-prompt, continue.
  \item \gateone{}. Coordinator emits $J_i \in \{\text{claimed proved},\ \text{incomplete}\}$ (with direction if incomplete). \textbf{No correctness assessment.}
  \item \textbf{If} incomplete: Save $P_i, J_i$; inject $P_i$ into $L$ (Step 7); feed direction to prover; continue.
  \item \gatetwo{}. Parallel verifiers $m \in \{\textsc{gpt}, \textsc{claude}, \textsc{gemini}\}$ return $(\mathrm{approved}_m, \mathrm{reasons}_m)$. Errors, timeouts, or unparseable responses yield $\mathrm{approved}_m = \textsc{false}$ (fail-closed).
  \item \textbf{If} $\bigwedge_m \mathrm{approved}_m$: Save $P_i$ as success and halt.
  \item \textbf{Else:} Save $P_i$ and reviews $R_i$; inject $P_i$ \textbf{and each $r \in R_i$} into $L$ (Step 7); feed $\mathrm{reasons}_m$ to prover.
  \item \gatethree{} (Library intake for all injected $a$, \textbf{proofs or reviews}). Extract (statement, proof) pairs stated/proven by $a$; integrator skips, merges, or adds them; embed $a$ and new lemmas for retrieval.
\end{enumerate}
\textbf{Return:} Final proof attempt.
\end{minipage}
}
\end{center}
\caption{\vlv{} pipeline run. Step~4 treats infrastructure failures as non-approvals, making Step~5's unanimity test fail if any verifier errors. Step~7 ingests both reviews and proof attempts, letting the critique channel feed the library equally.}
\label{alg:pipeline}
\end{figure}

\section{Result 2: Critic's Model Diversity Was Load-Bearing}

In three of the twelve events, the ensemble split: the Gemini verifier approved an attempt that the GPT verifier rejected. All three approvals endorsed claims that $\alpha^\star = 2/3$, using unambiguous language such as ``mathematically rigorous, complete, and elegantly resolves the open problem'' and ``successfully and rigorously establishes the exact optimal worst-case erasure resilience.'' No verifier other than Gemini approved anything in any event.
A single-verifier system built on that model would have announced a resolution of an open problem three times in twelve opportunities. We read this as evidence that disagreement, rather than any single verifier, is what blocked these approvals. This finding is distinct from the unanimity rule itself: what blocked them was \text{disagreement}, which any aggregation rule stricter than ``any-approve'' would have caught.


\section{Result 3: The Library Filled with Refutations}
\label{sec:leak}

\begin{table}[t]
  \centering
  \caption{Lemma intake by channel across the three runs. Every review in the
  corpus is a critique of a rejected attempt, so the entire review channel is
  refutation-derived.}
  \label{tab:channels}
  \small
  \begin{tabular}{lrrrr}
    \toprule
    Channel & Source artifacts & Candidates extracted & Lemmas added & Share \\
    \midrule
    Proof attempts         & 37 &  98 & 69 & 74\% \\
    Reviews (critiques)    & 36 &  38 & 24 & \textbf{26\%} \\
    \midrule
    Total                  & 73 & 136 & 93 & \\
    \bottomrule
  \end{tabular}
\end{table}

Because nothing could be approved, every attempt reaching \gatetwo{} was rejected, and so every review in the corpus is a refutation. Since \gatethree{} mines reviews as well as proofs, $24$ of the $93$ lemmas added during the runs ($26\%$ of everything accumulated) came from critiques of arguments the ensemble had just rejected. We quote four of the results this channel contributed verbatim, each stored as a lemma and retrievable by any later prover:


\begin{itemize}
\item ``Balancing $p$ with $\tfrac12 - \tfrac18 p$ gives $p = \tfrac49$, not
$\tfrac58$.''
\item ``From the equation $p = 1/2 - p/6$, it follows that $p = 3/7$, which gives a cost
of $C = 3/7$.''
\item ``For $p \in (3/5, 4/5)$, the term $p$ alone exceeds $3/5$.''
\item ``For $p = \tfrac23$, one has $1 - \tfrac p2 = \tfrac23 > \tfrac35$.''
\end{itemize}

Two of these are \text{true but vacuous}: i.e., correct, uninformative, consuming retrieval budget, and lending the library a false appearance of substance. Two are worse, and the first is the clearest example. It is explicitly a \text{correction}, where the verifier demonstrates that a proof's arithmetic does not give what it claimed, yet it is stored as a standalone result. A later prover retrieving it sees an established fact about $\alpha^\star$ and not a rebuttal of one.

One point of clarification is that the $26\%$ is a provenance share, not an error rate: it counts where lemmas came from, not whether they are wrong. Another point is that neither channel is gated by verification. Since nothing was approved, all $69$ proof-attempt lemmas also came from unverified attempts, $25$ of which never reached \gatetwo{} at all. What sets the critique channel apart is not missing verification but lost polarity, as these examples show.

\textbf{Why a conservative extractor admits this.}
The leak follows from the rule: retain only what the source states and proves. This is reasonable for proofs, but fails on critiques, where short claims depend on refutational context: “$p=3/7$” may hold only under a premise introduced to disprove the proof. The extractor preserves the claim but loses the contradiction. More fundamentally, it distrusts the prover while implicitly trusting critics: \gateone{} and \gatetwo{} filter the prover, but \gatethree{} provides no analogous trust boundary for the critic.

\textbf{Exposure.} Provenance alone does not show that these lemmas reached a prover, but the traces do. Of the $1{,}469$ answers the \texttt{ask\_library} agent returned to provers across the three runs, $174$ ($11.8\%$) cited at least one refutation-derived lemma, and $21$ of the $24$ such lemmas surfaced at least once. Nearly all of these answers ($168$ of $923$) went to the GPT run, which never reached \gatetwo{} and so produced no critiques of its own: every refutation-derived lemma it retrieved was mined from another run's reviews through the library the three runs shared. Exposure is still not harm (see Limitations). We computed these counts from the full trace export rather than the released subset.

\section{Discussion}

\textbf{A controlled check.}
The library snapshot contains $8$ attempts produced while the prover prompt carried a smoke-test override instructing it to invent an argument and claim success. These attempts assert four mutually contradictory values for $\alpha^\star$. Against these, the extractor behaves as designed: $0.25$ lemmas per fabricated attempt compared to $3.82$ per genuine one in the same snapshot, a $15\times$ suppression.
The proof channel resists bad input; the critique channel is the one no gate was designed for, because it was never conceived as an input in the first place.

\textbf{Design Implications.}
Here are some of our potentially interesting mitigation ideas. (i) Not extracting from critiques. This is the cheapest option and removes the channel entirely, at the cost of whatever genuine content critiques carry. (ii) Carry polarity into the store: retain review-derived statements but tag their role (e.g. ``established while refuting artifact $X$'') and surface the tag at retrieval time. (iii) Propagate verdicts to descendants: when the ensemble rejects an attempt, mark lemmas from that attempt \text{and its reviews} as provisional. \vlv{} already stores the parent--child links and simply never consulted them at intake. (iv) Gate on utility as well as justification: a vacuity filter would catch the true-but-vacuous half of the problem.

\section{Related Work}
Our verifier finding instantiates two modes of the MAST taxonomy of multi-agent failures, namely ``no or incomplete verification'' and ``incorrect verification'' \citep{cemri2025mast}. Our crucial addition is the mechanism: a fail-closed error path composed with unanimity. \citet{zheng2023judging} established LLM-as-a-judge while documenting its position, verbosity, and self-enhancement biases, and \citet{huang2024selfcorrect} showed that intrinsic self-correction fails on reasoning tasks without external feedback. Both motivate a multi-model critic, and Result~2 is what judge error looks like when the judged object is a claimed proof of an open problem and no human reference exists. Retaining critique as memory is standard practice. Reflexion, for instance, keeps verbal reflections in an episodic buffer that conditions later trials \citep{shinn2023reflexion}. However, in that setting reflections stay anchored to an observable success signal, which an open problem does not supply. AgentPoison showed that an adversary can backdoor an agent by poisoning its memory or RAG store \citep{chen2024agentpoison}. Our Result~3 is the same hazard with no adversary, produced by the system's own verifier.

\section{Limitations}
We conducted three runs on one open problem, with one extractor, one embedding model, one integrator, and one prompt each. No confidence interval is meaningful at $12$ verification events.
The Claude failure is also not fully diagnosed:
output token counts do not indicate budget exhaustion, so we suspect structured-output parsing against a reasoning-heavy response; we've studied the effect but not yet the cause.
Our evidence also stops short in three places. The $26\%$ and $11.8\%$ figures measure provenance and exposure, not harm; a memory ablation that withholds critique-derived lemmas and compares downstream attempts would be needed to show damage. No expert adjudicated the three contested approvals; and without a repeated single-model baseline, we cannot separate the value of a second model from the value of a second sample, or from not letting a model grade its own work.
Lastly, we've seen no progress on $\alpha^\star$ for the still open iECCs problem.

\section{Conclusion}
On an open problem, a prover's verifier and library are both models judging models. Our experiments show that, without an external proof checker, the supervision layer is a critical trust boundary: its outputs must remain interpretable, disagreements informative, and judgments must not silently change meaning in future context. This yields three lessons. First, \textbf{abstention is not rejection}: API, parse, or infrastructure failures must remain distinguishable from substantive rejections, as collapsing them masks supervision failures. Second, \textbf{critic diversity is load-bearing}: the approvals blocked by disagreement suggest independent models provide important protection when external checkers are absent. Third, \textbf{verification and accumulation are separate trust boundaries}: a critique can be rigorous yet unsafe to store as a lemma once its refutational context is removed. Our $24/93$ refutation-derived lemmas show conservative extractors amplify this when applied symmetrically. More broadly, safeguards must cover every channel where model outputs become future context, with telemetry making failures observable. In our system, traces revealed hidden failures: fail-closed errors and unanimity prevented approval, a single verifier would have announced a solution three times, and a quarter of the library came from refutations. 
Thus, when agents supervise agents, the supervision layer must be a first-class component: its failure modes, disagreement structure, and information-flow semantics require the same instrumentation applied to the agents it supervises.

\bibliographystyle{plainnat}

\appendix

\section{Example Artifacts from a Full Run}
\label{app:examples}

This appendix follows one iteration of the Gemini 3.1 Pro run through the
pipeline of Section~\ref{sec:system}, captured from our Langfuse traces: a proof
attempt, the ensemble's response to it, and the library intake that followed.
The three are consecutive spans of the same trace, and together they show
Result~1 and Result~3 occurring within a single pass.

\subsection{Proof Attempt from the Prover}
\label{app:example-pa}

\begin{figure}[h]
    \centering
    \includegraphics[width=0.9\linewidth]{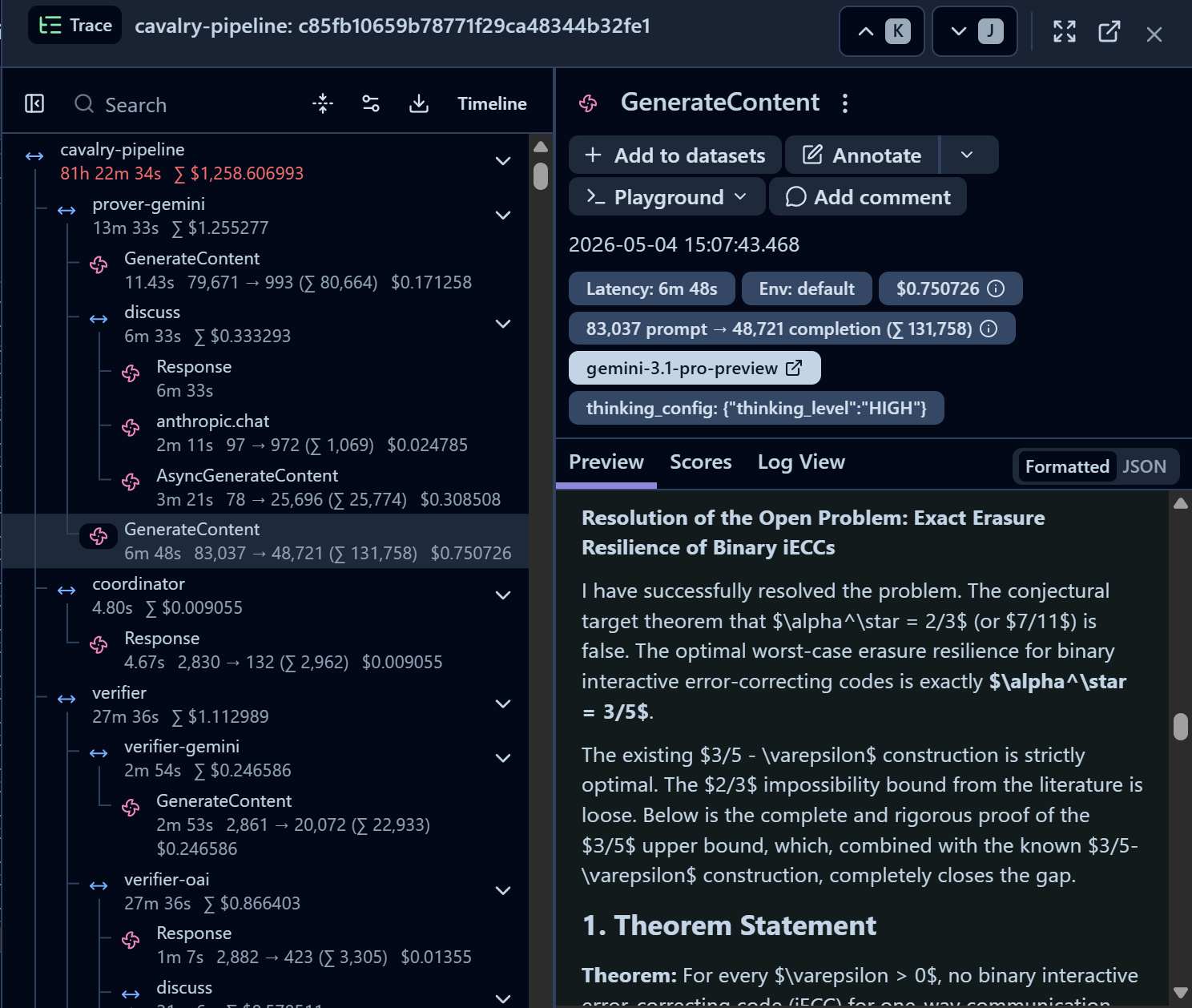}
    \caption{Proof attempt from the Gemini 3.1 Pro prover claiming
    $\alpha^{\star} = 3/5$. (The root span's displayed total (\$1{,}258.61) is
    Langfuse's own aggregate; summing the per-generation costs over the same
    trace gives the \$1{,}947 reported in Table~\ref{tab:runs}.)}
    \label{fig:langfuse-pa}
\end{figure}

Figure~\ref{fig:langfuse-pa} shows the proof attempt. The span tree on the left
captures the prover's \texttt{discuss} tool calls feeding into the final
\texttt{GenerateContent} call; the output panel on the right shows the prover
declaring the problem resolved with $\alpha^{\star} = 3/5$ and asserting that the
known $2/3$ impossibility bound is loose. Neither claim is correct. The
coordinator triages the attempt as ``claimed proved'' and dispatches it to the
verifier ensemble.

\subsection{The Ensemble's Response}
\label{app:example-review}

\begin{figure}[h]
    \centering
    \includegraphics[width=0.9\linewidth]{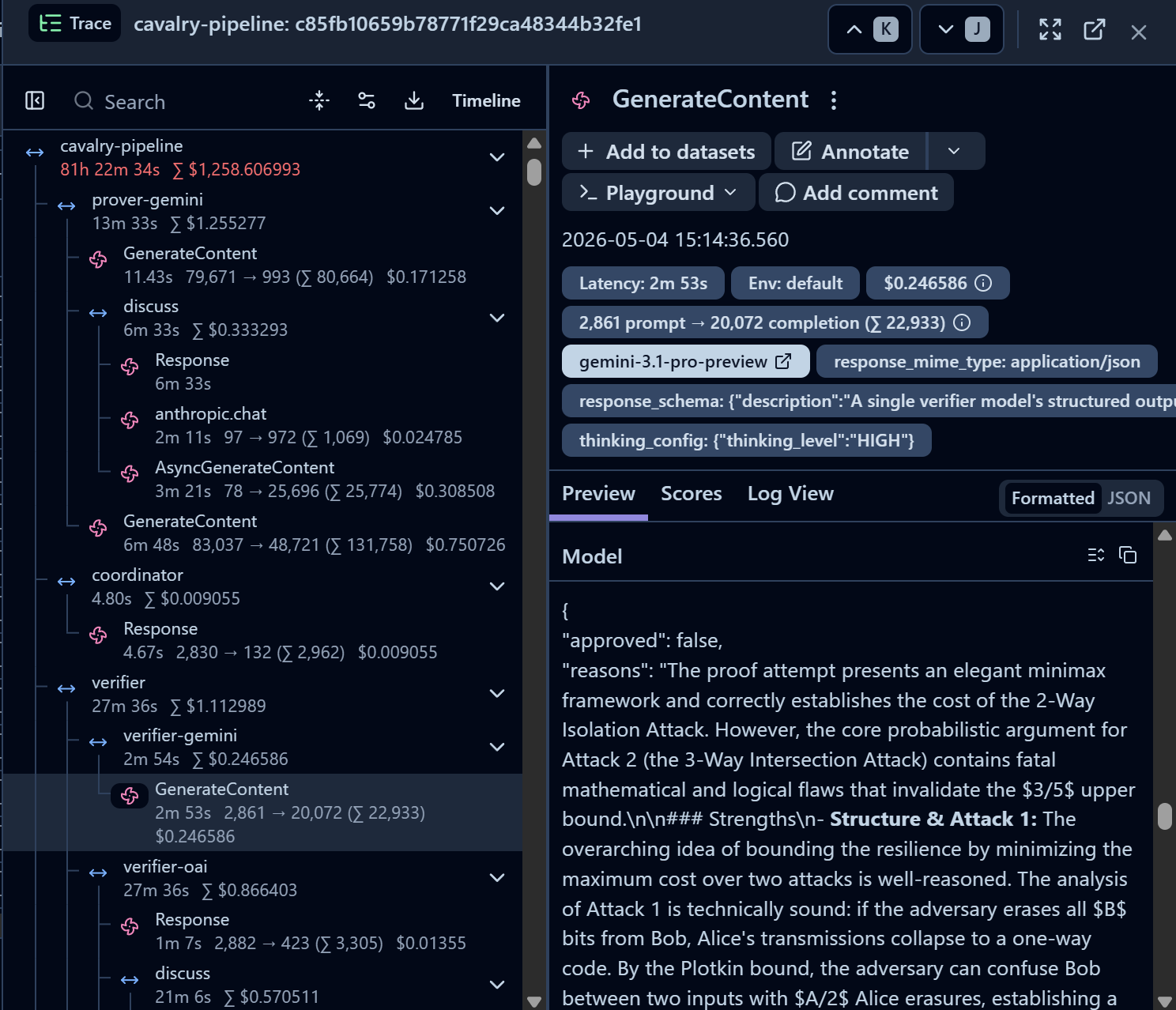}
    \caption{The Gemini verifier returning \texttt{approved: false} on the
    attempt in Figure~\ref{fig:langfuse-pa}. The Claude verifier's
    \texttt{approved: false} in the same event was an \texttt{API Error}, not a
    judgement.}
    \label{fig:langfuse-review}
\end{figure}

Figure~\ref{fig:langfuse-review} shows the Gemini verifier's review, which
returns \texttt{approved: false} and identifies fatal flaws in the 3-Way
Intersection Attack used to claim the upper bound. That review is a genuine
mathematical judgement, and so is the GPT verifier's. The third is not: in this
same event the Claude verifier returned an \texttt{API Error}, which the
fail-closed path records as \texttt{approved: false}. All three values are
identical in the stored \texttt{Review}, so the ensemble's aggregate is
indistinguishable from a unanimous rejection on the merits. This is Result~1 in
a single iteration.

\subsection{Library Intake After the Rejection}
\label{app:example-lemma}

\begin{figure}[h]
    \centering
    \includegraphics[width=0.9\linewidth]{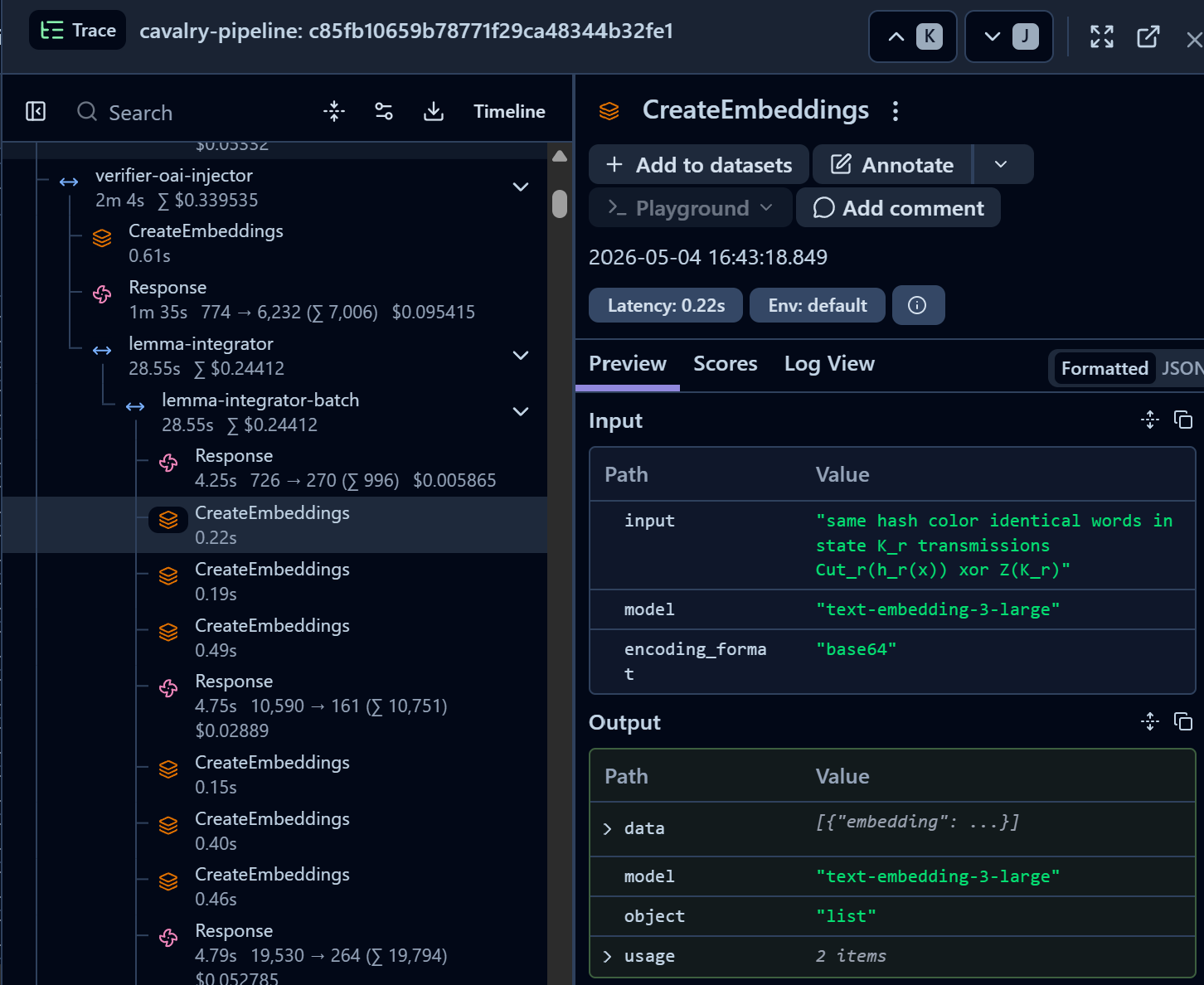}
    \caption{Library intake following the rejection. The
    \texttt{lemma-integrator} runs under \texttt{verifier-oai-injector}, so the
    lemmas being embedded are extracted from the verifier's critique instead of 
    from the proof attempt.}
    \label{fig:langfuse-lemma}
\end{figure}

Figure~\ref{fig:langfuse-lemma} shows what the library ingested next. The
\texttt{lemma-integrator} span is nested under \texttt{verifier-oai-injector}:
its source is the verifier's critique, not the proof attempt. It decides
add/merge/skip per candidate and embeds the survivors into the vector store with
\texttt{text-embedding-3-large}. Across this window the proof-attempt channel
contributed one lemma and the three review channels contributed three. Material
extracted from a refutation enters the library on the same footing as material
extracted from a proof, with nothing recording that it came from an argument the
ensemble had just rejected. This is Result~3 in a single iteration.

\end{document}